\documentclass[letterpaper, 10 pt, conference]{ieeeconf}
\usepackage{amssymb,mathrsfs,mathtools,amsmath,dsfont}
\usepackage{moreverb}
\usepackage{bm}
\usepackage{alltt}
\usepackage{makeidx}
\usepackage{multirow}
\usepackage{multicol}
\usepackage{cite}
\usepackage[dvipsnames,svgnames,table]{xcolor}
\usepackage{graphicx}
\usepackage{ulem}
\usepackage{hyperref}
\usepackage[english]{babel}
\usepackage[utf8]{inputenc}
\usepackage{fancyhdr}
\usepackage{color,soul}
\usepackage[utf8]{inputenc}
\usepackage{amsfonts}

\newtheorem{remark}{Remark}

\title{\LARGE \bf On the Controllers Based on Time Delay Estimation for	Robotic Manipulators}
\author{M. Reza J. Harandi
	\thanks{\textbf{A}dvanced
		\textbf{R}obotics and \textbf{A}utomated \textbf{S}ystems (ARAS),
		Faculty of Electrical Engineering,
		K. N. Toosi University of Technology,
		Tehran, lran.
		{\tt\small jafari@email.kntu.ac.ir} }
}	
\begin{document}

	\maketitle
	\thispagestyle{empty}
	\pagestyle{empty}
	\begin{abstract}
		Assurance of asymptotic trajectory tracking in robotic manipulators with a smooth control law in the presence of unmodeled dynamics or external disturbance is a challenging problem. Recently, it is asserted that it is achieved via a rigorous proof by designing a traditional model-free controller together with time delay estimation (TDE) such that neither dynamical parameters nor conservative assumptions on external disturbance are required. The purpose of this note is to show that this claim is not true and the stability proof of the method is incorrect. Finally, some modified versions of this controller with rigorous proof is presented for robotic manipulators.  
	\end{abstract}
	
	
	\section{Introduction}\label{s1}
	The performance of a robotic system is subject to some common
	challenges such as unmodeled dynamics, external disturbance,
	parameter uncertainties, etc. Several methods have been proposed to address these problems, and as a representative, adaptive controllers to address parametric uncertainties
	\cite{namvar2013adaptive,zamani2020distributed,harandi2021adaptive1,sharifi2021impedance,franco2021ida} and robust controllers to reject unmodeled dynamics/ external disturbance~\cite{franco2020robust,rahimian2013structural,karami2019smooth,shariati2016descriptor}
	could be listed. In adaptive control, it is usually assumed that the
	structure of the model is known, but the parameters are uncertain.
	Robust controllers can reject special types of disturbance, while some of them suffering from chattering in response. Under these assumptions and cumbersome calculations in some methods, asymptotic trajectory tracking is mathematically ensured.
	
	Recently, it has been claimed that the design of a simple model-free
	controller with time delay estimation (TDE) makes it possible to
	guarantee trajectory tracking; see for
	example~\cite{zhang2021attitude,hosseini2021practical,wang2020model,kali2018super,wang2019new,zhang2020fractional} and references therein. In this
	method, the robot's dynamic model is not required to be known
	in the controller synthesis, but instead, the dynamic behavior of
	the system in the previous time steps is used. By this means, a
	desired model with a constant inertia matrix is considered, and a
	simple controller synthesis without knowing the system dynamics is performed. Hereupon, it is claimed that 
	trajectory tracking is ensured mathematically. For this purpose, it is asserted in their proof (see for example, Theorem~1 of \cite{wang2020model,kali2018super,wang2019new} and section~5 of \cite{zhang2020fractional}) that the dynamic of TDE error is represented by a set of linear discrete-time equations with an Hurwitz matrix and thus, the TDE error has an upper bound. A claim which is totally questionable. 
	
	In this note, the stability of TDE-based controllers for robotic manipulators is investigated. 
	It is shown that the upper bound of TDE error is
	limited only if the stability of the manipulator is ensured. Hence,
	the upper bound of the TDE error is state-dependent and it is not independent of the system’s stability. Furthermore, the dynamic of TDE error is represented by a set of discrete-time linear time-varying equations and thus, its stability may not be achieved by merely Hurwitz condition~\cite[ch.~4]{khalil2002nonlinear}.
	Therefore, the stability of TDE-based controllers for robotic manipulators is doubtful.

	\section{TDE-based Controller Synthesis for Robotic Manipulators}\label{s2}
	Here, the main concept of TDE-based controllers for robotic systems is proposed.
	Consider dynamical formulation  of a $n$-DOF robot in the following
	form~\cite{wang2019new,kali2018super,wang2020adaptive1}
	\begin{align}
	M(q)\ddot{q}+C(q,\dot{q})\dot{q}+g(q)+f(q,\dot{q})+d=\tau,
	\label{1}
	\end{align}
	in which $q\in\mathbb{R}^n$ is joint position, $\dot{q}\in\mathbb{R}^n$ denotes velocity and $\ddot{q}\in\mathbb{R}^n$ is acceleration.
	$M(q)\in\mathbb{R}^{n\times
		n}$ and $C(q,\dot{q})\in\mathbb{R}^{n\times n}$ are the positive definite inertia matrix, and  the centrifugal
	and Coriolis matrix, respectively,
	$g(q)\in\mathbb{R}^n$ is the vector of gravity terms,
	$f(q,\dot{q})\in\mathbb{R}^n$ denotes the system natural damping
	terms, and $d,\tau\in\mathbb{R}^n$ denotes the external disturbance
	and control input, respectively. Note that in some papers,
	e.g.,~\cite{wang2019new}, the dynamic of the actuators is also
	considered. However, for simplicity, it is not considered in this note.
	Dynamic model (\ref{1}) is rewritten in the following form
	\begin{align}
	\overline{M}\ddot{q}+h(q,\dot{q},\ddot{q})=\tau,
	\end{align}
	where $\overline{M}\in\mathbb{R}^{n\times n}$ is a constant matrix.
	The vector $h$ that describes the remaining terms in the dynamic
	formulation is given by:
	\begin{align}
	h(q,\dot{q},\ddot{q})=(M-\overline{M})\ddot{q}+C(q,\dot{q})\dot{q}+g(q)+f(q,\dot{q})+d.
	\end{align}
	The TDE--based control law is designed as follows
	\begin{align}
	\tau=\overline{M}u+\hat{h},
	\end{align}
	in which $u$ is the main controller that is chiefly different types
	of sliding mode controller and $\hat{h}$ is estimated value of $h$
	which is derived base on TDE method as follows
	\begin{align}
	\hat{h}(t)\approx h(t-\rho)=\tau(t-\rho)-\overline{M}\ddot{q}(t-\rho),
	\label{tde}
	\end{align}
	in which $\rho$ denotes the time delay. The last term in  (\ref{tde}) is usually unknown. Hence, the following
	numerical differentiation is used
	\begin{align*}
	\ddot{q}(t-\rho)\approx\big(q(t)-2q(t-\rho)+q(t-2\rho)\big)/\rho^2.
	\end{align*}
	In order to ensure trajectory tracking in the relevant articles, first, it has been shown
	that the TDE error and its derivatives are bounded, and then, the
	stability of the closed-loop system is guaranteed via the direct
	Lyapunov method. The critical part of the proof is to ensure the
	boundedness of ${h}-{\hat{h}}$ and its derivative. In the next
	section, this issue is analyzed in detail.
	\section{Main Results}
	\subsection{Analysis of TDE error}\label{s31}
	For the sake of simplicity, define
	$$e=\overline{M}^{-1}(\hat{h}-h).$$
	Using (\ref{1}-\ref{tde}), the following equations are derived
	\begin{align}
	&Me=M(\ddot{q}-u)=\tau-d-f-g-C\dot{q}-Mu\nonumber\\&=\big(\overline{M}-M\big)u+\hat{h}-\mathcal{N}(t)=\big(\overline{M}-M\big)u\nonumber\\
	&+\big(M(t-\rho)-\overline{M}\big)\ddot{q}(t-\rho)+\mathcal{N}(t-\rho)-\mathcal{N}(t),
	\label{me1}
	\end{align}
	with
	\begin{align*}
	\mathcal{N}(t)=d(t)+f(t)+g(t)+C(t)\dot{q}(t).
	\end{align*}
	Note that all the vectors and matrices are at time $t$
	unless indicated. Since $\ddot{q}(t-\rho)=e(t-\rho)+u(t-\rho)$,
	(\ref{me1}) is rewritten as follows
	\begin{align}
	Me&=(M-\overline{M})e(t-\rho)-(M-\overline{M})\big(u-u(t-\rho)\big)\nonumber\\&+\big(M(t-\rho)-M\big)\ddot{q}(t-\rho)+\mathcal{N}(t-\rho)-\mathcal{N}(t),
	\end{align}
	where $\pm M\ddot{q}(t-\rho) $ was added to the last equation of (\ref{me1}).
	Therefore, the error $e$ may be expressed in the following form
	\begin{align}
	\label{dis}
	e(t)=De(t-\rho)+\xi
	\end{align}
	with
	\begin{align}
	D&=I-M^{-1}(t)\overline{M}\nonumber\\
	\xi&=D\big(u-u(t-\rho)\big)+M^{-1}\Big(\big(M(t-\rho)-M\big)\ddot{q}(t-\rho)\nonumber\\&+\mathcal{N}(t-\rho)-\mathcal{N}(t)\Big).
	\label{xi}
	\end{align}
	Equation (\ref{dis}) may be interpreted as a discrete system with
	input $\xi$. In the literature, it is asserted that
	bv suitable design of $\overline{M}$, the system (\ref{dis}) is bounded-input bounded-output and therefore,
	TDE error is bounded if $\xi$ is bounded. Additionally, the upper
	bound of derivative of TDE error is given as:
	\begin{align}
	\|\dot{e}\|&\approx\|\frac{e(t)-t(t-\rho)}{\rho}\|=\frac{1}{\rho}\|-M^{-1}\overline{M}e(t-\rho)+\xi\|\nonumber\\&\leq\frac{1}{\rho}(\|M^{-1}\|\|\overline{M}\|\|e(t-\rho)\|+\|\xi\|).
	\end{align}
	Thus, the derivative of the TDE error is bounded if TDE error and
	$\|\xi\|$ are bounded. By considering (\ref{xi}), it is clear that
	$\xi$ depends on the position, velocity, and acceleration. Thus,
	boundedness of $\xi$ is only ensured if $q,\dot{q}$ and $\ddot{q}$
	are bounded. This means that for the boundedness of the TDE error, stability of the system is required. It is a common
	technical error in TDE-based papers, in which the cause and effect are
	intertwined. Additionally, the matrix $D$ given in (\ref{xi}) ic not clearly constant. Hereupon, the system (\ref{dis}) is linear time-varying and its stability is not deduced from Hurwitz condition~\cite[ch.~4]{khalil2002nonlinear} and should be analyzed via other methods such as Lyapunov.
	Hence, according to these reasons, it is inferred that the stability of TDE-based controllers is not correct, and
	the proposed experiments/ simulations in the related
	articles are not confirmed by mathematical proof.
	\begin{remark}\normalfont
		In some papers, such as~\cite{kali2018super,wang2019new} the
		authors have tried to address this problem through another
		unrealistic assumption. They have presumed that the velocity and
		acceleration are bounded. Although in practical implementations this
		might be observed; however, this assumption is equivalent to
		assuming the stability of the closed-loop system.
		Furthermore, it is clear that an assumption should be defined such
		that it is verifiable at least in simulation, e.g., type of external
		disturbance, initial conditions, the precise values of states, etc.
		The boundedness of states and their derivatives should be assured
		by suitable control law and rigorous stability analysis. Therefore,
		this assumption does not rectify the problem while it is an obvious
		contradiction with the aim of the stability analysis.
	\end{remark}
	\begin{remark}\normalfont
		As explained in  section~\ref{s2}, designing a TDE-based
		controller is based on a traditional controller together with a
		TDE-based component. Therefore, an extension of the method which
		could be applicable to a wide range of systems seems to be simple and is proposed
		as follows. As an example, consider interconnection and damping
		assignment passivity-based control (IDA-PBC) approach, which is a
		comprehensive method that stabilizes the general system
		$\dot{x}=f(x)+g(x)u$ at the equilibrium point $x^*$ through solving
		a set of partial differential equations
		(PDEs)~\cite{franco2020robust,harandi2021solution}.
		It may seem that combining IDA-PBC  with a TDE-based component,
		makes it possible to stabilize a system since the solutions of PDEs
		are selected freely and the remaining terms in the closed-loop
		dynamics resulted from the terms not satisfying the matching
		equations, could be considered as disturbance and rejected by TDE
		part. However, as explained before, the stability
		proof of TDE-based controllers is incorrect. Hence, the
		difficulty of solving a set of PDEs is the expense of precise stability assurance.
	\end{remark}

	\subsection{Overall stability analysis}
	Under the boundedness of TDE error, it is easy to ensure boundedness of the tracking error with a continuous control law, or asymptotic trajectory tracking with a non-continuous controller. However, in some of the papers that the main controller is based on higher order sliding mode such as \cite{wang2019new,kali2018super,wang2019adaptive,wang2019adaptive1}, it is argued that asymptotic stability is ensured via a continuous controller. Consider the case where the main controller is the super-twisting algorithm.
	It has been asserted that the closed-loop equations are expressed in the following form
	\begin{align}
	&\dot{s}=-K_1\Lambda(s)\text{sign}(s)+\Omega\nonumber\\
	&\dot{\Omega}=-K_2\mbox{sign}(s)+\dot{e}
	\label{mo}
	\end{align}
	with $\Lambda(s)=\mbox{diag}[|s_1|^{0.5},...,|s_n|^{0.5}]$ and $s$ is sliding surface.
	 Then by considering a Lyapunov function in the form $V=\eta^TP\eta$ with $\eta=[\Lambda(s)\text{sign}(s)^T,\Omega^T]^T$, it has been shown that the derivative of the Lyapunov function is in the form $-\eta^TQ\eta$ such that by suitable values of $K_1$ and $K_2$, the matrix $Q$ is positive definite with respect to boundedness of $\dot{e}$. Although it was shown that boundedness of $\dot{e}$ is not independent of stability,
	the model (\ref{mo}) is not correct since from (\ref{me}) it is clear that $\dot{e}$ is a function of $\ddot{q}$ which is not a state of a robotic system. The correct representation of (\ref{mo}) is in the following form
	\begin{align}
	&\dot{s}=-K_1\Lambda(s)\text{sign}(s)+\Omega+e\nonumber\\
	&\dot{\Omega}=-K_2\mbox{sign}(s)
	\label{moo}
	\end{align}
	By considering the same Lyapunov function, its derivative is $-\eta^TQ\eta$ while the matrix $Q$ is indefinite. This shows the stability proof is not correct.
	
	In the sequel, this method is modified and applied to a robotic
	system for some particular cases. Notice that the aim is designing
	TDE-based controllers with precise proof of stability, and then,
	compare that to the state-of-the-art controllers.
	
	\subsection{Design of TDE-based controller}\label{s33}
	\subsubsection{Case 1}
	Consider a robotic manipulator with the dynamic equation (\ref{1}).
	Assume that the dynamic parameters and $\ddot{q}(t-\rho)$ are known.
	Additionally, presume that $\|\dot{d}\|\leq \epsilon$ with
	$\epsilon$ being a positive known value. The control
	law which is the combination of Slotine-Li
	controller~\cite{slotine1989composite} and a TDE term is given as
	follows
	\begin{align}
	\label{tau}
	\tau=M\dot{\nu}+C\nu+g-KS+\hat{d}(t),
	\end{align}
	with
	\begin{align}
	\nu&=\dot{q}_d-\Gamma \tilde{q}, \qquad S=\dot{q}-\nu,
	\qquad\tilde{q}=q-q_d,\nonumber\\
	\hat{d}(t)&=d(t-\rho)=\tau(t-\rho)-M\ddot{q}(t-\rho)\nonumber\\
	&\hspace{1mm}-C(t-\rho)\dot{q}(t-\rho)-g(t-\rho),\label{dhat}
	\end{align}
	where $q_d\in\mathbb{R}^n$ is the desired trajectory and $\Gamma,K\in\mathbb{R}^{n\times n}$ are positive definite gains. Consider
	\begin{align}
	V=\frac{1}{2}S^TMS, \label{ly}
	\end{align}
	as a Lyapunov function candidate, it is straightforward to  compute
	the upper bound of its derivative, which is given as
	$$\dot{V}=-S^TKS-S^T\big(d(t)-\hat{d}(t)\big),$$
	in which the property that $\dot{M}-2C$ is skew-symmetric, was used.
	Since $\dot{d}$ is bounded, $d(t)-\hat{d}(t)$ is also bounded with
	the upper bound $\|d(t)-\hat{d}(t)\|\leq \rho\epsilon$. Thus
	$$\dot{V}\leq -S^TKS+\rho\epsilon\|S\|\leq -\|S\|
	(\lambda_{min}\{K\}\|S\|-\rho\epsilon),$$
	which shows that $\dot{V}\leq-\beta\|S\|^2$ if
	$$\|S\|\geq \frac{\rho \epsilon}{\lambda_{min}\{K\}-\beta},$$
	with $\beta$ being an arbitrary small value. Hence, $S$ and
	consequently $\tilde{q}$ have an ultimate bound. Since
	$$\frac{1}{2}\lambda_{min}\{M\}\|S\|^2\leq V\leq \frac{1}{2}\lambda_{max}\{M\}\|S\|^2,$$
	the ultimate bound of $\|S\|$ is derived as follows
	$$\|S\|\leq \sqrt{\frac{\lambda_{max}\{M\}}{\lambda_{min}\{M\}}}
	\frac{\rho \epsilon}{\lambda_{min}\{K\}-\beta}.$$
	Note that boundedness of error is ensured in the presence of
	(possibly unbounded) disturbance with the expense of the precise
	knowledge of $\ddot{q}(t-\rho)$. This is certainly one of the
	superiority of the TDE-based controller compared to other reported
	controllers in the literature. Notice that if $d$ is a constant
	external disturbance, the ultimate bound is replaced by asymptotic
	trajectory tracking.
	\subsubsection{Case 2}
	Now, consider a class of robotic systems in which $g(q)$ is
	bounded with the upper bound $\|g(q)\|\leq \kappa$. This is a very
	light and feasible assumption fully applicable in many case studies~\cite{harandi2021adaptive,kelly1997pd}.
	In this case, it is possible to compensate the gravity term $g(q)$
	using the TDE method. Therefore, the control law (\ref{tau}) is
	modified as follows
	\begin{align}
	\label{tau1}
	\tau&=M\dot{\nu}+C\nu-KS+\hat{h}(t),\\
	\hat{h}(t)&=h(t-\rho)=d(t-\rho)+g(t-\rho)=\tau(t-\rho)\nonumber\\
	&\hspace{17mm}-M\ddot{q}(t-\rho)-C(t-\rho)\dot{q}(t-\rho).\nonumber
	\end{align}
	Consider (\ref{ly}) as the Lyapunov function candidate, and derive
	its derivative as:
	\begin{align*}
	\dot{V}&=-S^TKS-S^T\big(g(t)-g(t-\rho)\big)-S^T\big(d(t)\\&-d(t-\rho)\big)
	\leq -\|s\|(\lambda_{min}\{K\}\|s\|-2\kappa-\rho\epsilon),
	\end{align*}
	where $\|g(q)-g(q-\rho)\|\leq 2\kappa$ was substituted. Hence,
	similar to previous case, the tracking error has an ultimate bound
	such that the upper bound $\|S\|$ is
	$$\|S\|\leq \sqrt{\frac{\lambda_{max}\{M\}}{\lambda_{min}\{M\}}}
	\frac{2\kappa+\rho \epsilon}{\lambda_{min}\{K\}-\beta}.$$
	Note that these two cases are applicable to regulate a robotic
	system without any knowledge about requirement Dynamic matrices $M$
	and $C$. In this situation, the control law (\ref{tau}) and (\ref{tau1})
	are modified respectively, as
	$$\tau= -K_p\tilde{q}-K_d\dot{q}+g+\hat{d},$$
	and
	$$ \tau=-K_p\tilde{q}-K_d\dot{q}+\hat{h},$$
	with $\hat{d}$ and $\hat{h}$ defined in (\ref{dhat}) and
	(\ref{tau1}), respectively, and $0<K_p,K_d\in\mathbb{R}^{n\times n}$. Consider
	$$V=\frac{1}{2}\dot{q}^TM\dot{q}+\frac{1}{2}\tilde{q}^TK_p\tilde{q},$$
	as a Lyapunov candidate, its derivative in the first case
	is
	\begin{align*}
	\dot{V}&=-\dot{q}^TK_d\dot{q}-\dot{q}^T\big(d(t)-\hat{d}(t)\big)
	\\&\leq -\|\dot{q}\|(\lambda_{min}\{K_d\}\|\dot{q}\|-\rho\epsilon),
	\end{align*}
	and in the second case is derived as
	\begin{align*}
	\dot{V}&=-\dot{q}^TK_d\dot{q}-\dot{q}^T\big(d(t)-\hat{d}(t)\big)-
	\dot{q}^T\big(g(t)-g(t-\rho)\big)\\&\leq -\|\dot{q}\|(\lambda_{min}\{K_d\}
	\|\dot{q}\|-\rho\epsilon-2\kappa),
	\end{align*}
	which show that the error has an ultimate bound.
	
	Note that in the cases where the inertia matrix or the centrifugal
	and Coriolis matrix are
	unknown, as explained in section~\ref{s2}, it is not possible to
	ensure the system's stability with a TDE-based controller since
	the upper bound of the TDE error is related to velocity and acceleration
	of the system which are bounded if the manipulator is stable. 
	
	\subsubsection{Case 3}
	The results of  Case 1 are applicable to the problem of
	stabilization of port Hamiltonian (PH) systems with matched
	disturbance. Invoking~\cite{ferguson2020matched}, consider the
	following PH system
	\begin{align}
	\dot{x}=[J(x)-R(x)]\nabla_x H+G(x)\big(u-d(t)\big),
	\label{sys}
	\end{align}
	where $H$ denotes Hamiltonian, $J=-J^T\in\mathbb{R}^{n\times n}$ is
	interconnection matrix, $0\leq R(x)\in\mathbb{R}^{n\times n}$ is
	damping matrix, $G\in\mathbb{R}^{n\times m}$ is the input mapping
	matrix and $d$ denotes external disturbance such that
	$\|\dot{d}\|\leq \epsilon$. Assume that 
	$${x}^{*}=\text{{arg min}}\, H({x}).$$ 
	Note that (\ref{sys}) may represent the closed-loop equation of
	nonlinear input--affine systems such as underactuated robots, see
	\cite{ortega2004interconnection} for more details. Under the
	assumption that all the terms in (\ref{sys}) are known except the
	disturbance, the control law is given as 
	\begin{align*}
	u&=-KG^T\nabla_x H+\hat{d}(t)\\
	\hat{d}(t)&=d(t-\rho)=G^\dagger(t-\rho)\big(-\dot{x}(t-\rho)
	+[J(t-\rho)\\&-R(t-\rho)]\nabla_x H(t-\rho)\big)+u(t-\rho),
	\end{align*}
	where $K\in\mathbb{R}^{m\times m}$ is positive definite gain, and
	$G^\dagger$ denoted the left pseudo-inverse of $G$. Consider $H$ as
	the Lyapunov function, and find its derivative as
	\begin{align*}
	\dot{H}&=-(\nabla_x H)^T \Big((R+GKG^T)\nabla_x
	H+Gd(t)\\&-Gd(t-\rho)\Big)\leq -(\nabla_x H)^TR\nabla_x
	H-\|(\nabla_x H)^T
	G\|\\&\hspace{5mm}\big(\lambda_{min}\{K\}\|(\nabla_x H)^T
	G\|-\rho\epsilon\big),
	\end{align*}
	which shows that $x-x^*$ has an ultimate bound. Clearly, if $d$ is
	constant, then $x^*$ is (asymptotically) stable. In comparison to
	\cite{ferguson2020matched}, the advantage of the proposed TDE-based
	controller is its independence to disturbance dynamics, and the
	disadvantage is considering a particular disturbance and requirement
	to know $\dot{x}(t-\rho)$.
	
	Based on the proposed cases, we may deduce that TDE-based
	controllers are practical to reject particular types of external
	disturbance if dynamics of the system and also $\ddot{q}(t-\rho)$
	are known. Otherwise, they are not outperforming the
	state-of-the-art adaptive and robust controllers developed in
	the literature.
	
	\section{Conclusion}
	In this note, TDE-based controllers for robotic systems
	were analyzed. It was elaborated that the stability proof of this method is wrong since the upper bound of TDE error is a function of the position, velocity and acceleration of the system and thus, it is bounded if system's stability is previously ensured. Furthermore, due to the representation of TDE error by a set of discrete-time linear time-varying equations, Hurwitz condition does not imply the stability.

	\bibliographystyle{ieeetr}
	
	\bibliography{ref}
\end{document}